\documentclass{article}
\usepackage{spconf,amsmath,epsfig}
\usepackage{multirow}
\usepackage{graphicx}
\usepackage{times}
\usepackage{epsfig}
\usepackage{amsmath,fge}
\usepackage{amssymb}
\usepackage{subcaption}
\usepackage{euscript}
\usepackage{rotating}
\usepackage{bm}
\usepackage{notoccite}

\begin{document}\sloppy

\def\x{{\mathbf x}}
\def\L{{\cal L}}

\title{A Multi-Layer Approach to Superpixel-based Higher-order Conditional Random Field for Semantic Image Segmentation}
%
 \name{Li Sulimowicz, Ishfaq Ahmad, Alexander Aved}
 \address{Department of Computer Science and Engineering, University of Texas at Arlington, TX, USA, Rome Airforce Research Lab\\
 \textit{li.yin@mavs.uta.edu}, \textit{iahmad@cse.uta.edu}, \textit{alexander.aved@us.af.mil}}
\name{Li Sulimowicz$^{\star}$ \qquad Ishfaq Ahmad$^{\star}$ \qquad Alexander Aved$^{\dagger}$}
\address{$^{\star}$ Department of Computer Science and Engineering, University of Texas at Arlington, TX, USA\\
     $^{\dagger}$ Air Force Research Laboratory, Rome, NY, USA\\
     \textit{\{li.yin@mavs, iahmad@cse\}.uta.edu}$^{\star}$, \textit{alexander.aved@us.af.mil}$^{\dagger}$}





%

\maketitle
\begin{abstract}
Superpixel-based Higher-order Conditional random fields (SP-HO-CRFs) are known for their effectiveness in enforcing both short and long spatial contiguity for pixelwise labelling in computer vision.  However, their higher-order potentials are usually too complex to learn and often incur a high computational cost in performing inference. We propose an new approximation approach to SP-HO-CRFs that resolves these problems. Our approach is a multi-layer CRF framework that inherits the simplicity from pairwise CRFs by formulating both the higher-order and pairwise cues into the same pairwise potentials in the first layer. Essentially, this approach provides accuracy enhancement on the basis of pairwise CRFs without training by reusing their pre-trained parameters and/or weights. The proposed multi-layer approach performs especially well in delineating the boundary details (boarders) of object categories such as ``trees'' and ``bushes''.  Multiple sets of experiments conducted on dataset MSRC-$21$ and PASCAL VOC $2012$ validate the effectiveness and efficiency of the proposed methods. 
\end{abstract}

\section{Introduction}
\label{sec:intro}
Over the last few years, convolutional neural networks (CNNs) have been highly successful in a variety of computer vision tasks such as image recognition and semantic segmentation. However,  due to the loss of resolution and position information, CNNs always produce ``blobby'' output with coarse boundaries. Thus, probabilistic graphical models such as conditional random fields (CRFs) and higher-order CRFs (HO-CRFs) have been widely used together with CNNs~\cite{arnabconditional,Zheng2015cu, chen2016deeplab, arnab2016higher,chandra2016fast} for semantic image segmentation to enforce smoothness constraints for converting the output of CNNs from ``blobby'' to ``sharp.'' Specifically, HO-CRFs~\cite{kohli2009robust,komodakis2009beyond,vineet2014filter, arnab2016higher} have been demonstrated to be highly effective in enforcing long range or even global connections; for instance, region-level appearance consistency, co-occurrence of objects, and global connectivity. 

We observed two main problems with the traditional region-based HO-CRFs (Or superpixel-based HO-CRFs); a superpixel is a group of pixels with high appearance similarity and generated by segmentation algorithms~\cite{li2012segmentation,sulimowicz2017rapid}. First, the higher-order terms which are defined over cliques consisting of more than two nodes are computationally complex in both learning and inference process. For instance, transformation based methods, mean-field approximation, and dual decomposition. Second, the traditional higher-order terms~\cite{arnab2016higher,kohli2009robust} need to be learned on large training set, which requires a long training time before obtaining  effective and workable higher-order potentials. 
\setlength{\intextsep}{2pt}
\begin{figure}
\centering 
        
        \begin{subfigure}[b]{0.33\columnwidth}
                \includegraphics[width=.98\columnwidth, height = 2cm]{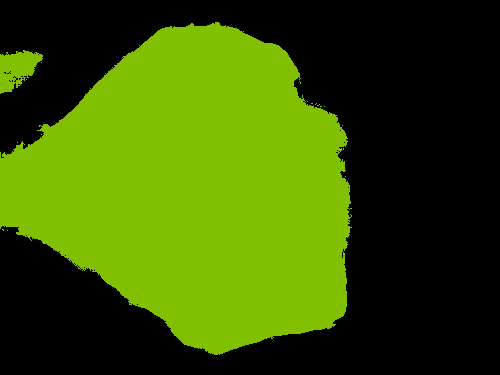}
                \caption{CRF-RNN~\cite{Zheng2015cu}}
        \end{subfigure}%
              \begin{subfigure}[b]{0.33\columnwidth}
                \includegraphics[width=.98\columnwidth, height = 2cm]{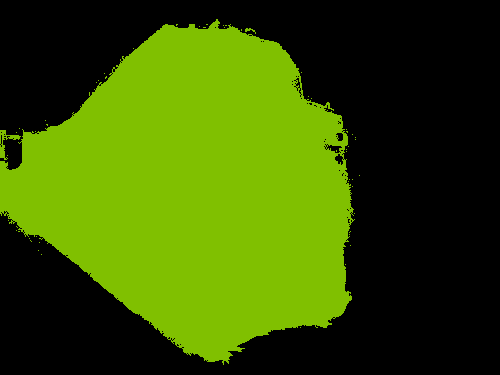}
                \caption{The proposed}
        \end{subfigure}%
                      \begin{subfigure}[b]{0.33\columnwidth}
                \includegraphics[width=.98\columnwidth, height = 2cm]{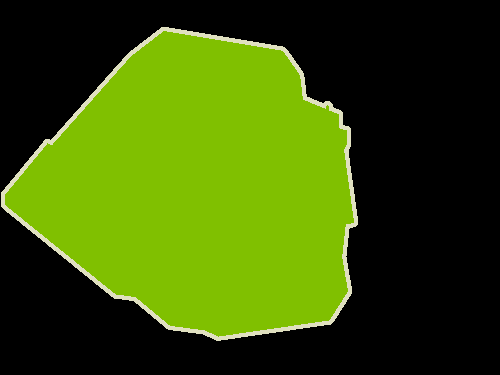}
                \caption{Ground Truth}
        \end{subfigure}%
              \caption{One example of the comparative result of the proposed method with CRF-RNN, where the proposed improves accuracy by getting rid of small spurious regions.}
        \label{fig:visual_result}
\end{figure}

To resolve these two problems, we propose an approximation method with a multi-layer framework, which has two layers and each layer itself is a pairwise CRF. Essentially, the first layer is called Segmentation as Input (SaI), which works by utilizing the color-sensitive feature of most pairwise potentials. SaI uses a segmented image, wherein each pixel takes the averaged RGB value of the superpixel it belongs to. SaI combines the superpixel-based cues into pairwise potentials and at the same partially preserves the pixelwise regulation. As a result, SaI itself can work alone as SP-HO-CRF. SaI provides the following two benefits: 1) decrease the computational complexity to the same level as that of pairwise CRFs. 2) reuse the parameters and/or weights from pre-trained pairwise CRFs, so that no training or just simple grid search is required to boost the accuracy. However, SaI slightly suffers from the pairwise potential information loss. The multi-layer framework with a pairwise CRF as the second layer compensates for this loss and further boosts the accuracy. We call this approximation approach SaI-based Multi-layer CRF (SM-CRF). SM-CRF preserves accurate boundaries for meshy objects such as trees and bushes. We applied SaI and SM-CRF on DenseCRF~\cite{Krahenbuhl2012tu} and CRF-RNN~\cite{Zheng2015cu}, and evaluated them using the dataset MSRC-$21$ and PASCAL VOC $2012$~\cite{hariharan2014simultaneous}, respectively. The experimental results show that SaI decreases the error rate by up to $16\%$ compared to the baseline models with only simple grid search training. One example is shown in Fig.~\ref{fig:visual_result}.

This paper is organized as: Section~\ref{sec:preliminaries} provides preliminaries related to pairwise CRFs and SP-HO-CRFs. Section~\ref{sec:proposed} gives details of our two main contributions, SaI and SM-CRF. Then, the experiments and related results are given in Section~\ref{sec:result}. Finally, we provide concluding remarks in Section~\ref{sec:conclusion}.
\section{Preliminaries}
\label{sec:preliminaries}
We start with the preliminaries about CRFs and Higher-order CRFs and introduce the notations used in this paper. 
\subsection{Conditional Random Fields~\cite{lafferty2001conditional}}
\label{definition}
Define two random fields $\bm{D}$ and $\bm{X}$. $\bm{D}$ is a vector of random variables $\{{D}_1, ... , {D}_N\}$, wherein ${D}_i$ is associated to pixel $i$ in the observed image. $\bm{X}$ is a vector of random variables over $\bm{D}$ and take value from a pre-defined label set $\EuScript{L} = \{l_1, ..., l_L\}$. Let $\EuScript{G}=(\EuScript{V},\EuScript{ E})$ be a graph over $\bm{X}$. The conditional random field $(\bm{D}, \bm{X})$ is formulated by a Gibbs distribution $P(\bm{X}|\bm{D})$=${\frac{1}{Z(\bm{D})}}{\text{exp}(-E(\bm{X}|\bm{D})}$. $E(\bm{X}|\bm{D})=\sum_{c\in \EuScript{C}_{g}}{\phi_c{(X_c|\bm{D})}}$ is called Gibbs energy and $Z(\bm{D})$ is the normalization constant known as partition function. Here $\EuScript{C}_{g}$ is a set of cliques in $\EuScript{G}$. The maximum a posteriori labeling of the random field is the same as $x^\ast = \arg\min_{x\in\mathcal{L}^{N}}E(\bm{X}|\bm{D})$.
\subsection{Superpixel-based Higher-order CRFs}
\label{pairwise_crf}
Conventionally, for superpixel-based higher-order CRFs~\cite{kohli2009robust, arnab2016higher}, 
the Gibbs energy is shown as follows: 
\begin{equation}
\label{ge_sp}
\resizebox{0.91\columnwidth}{!}
{
$E(\bm{X}|\bm{D})=\sum_{i\in \EuScript{V}}{\psi_i}^U{(x_i)}+\sum_{(i,j)\in\EuScript{E}}{\psi_{ij}}^P{(x_i,x_j)}+\sum_{s \in\EuScript{S} }{\psi_s}^{SP}{(X_s)}$
}
\end{equation}
here the Gibbs energy consists of sums of unary, pairwise and higher-order potentials.  $\EuScript{S}$ refers to a set of image segments.

\textbf{Gaussian Pairwise Potential in DenseCRFs~\cite{Krahenbuhl2012tu, chen2014semantic, Zheng2015cu, arnab2016higher,chen2016deeplab}} take the form as follows:
\begin{small}
\begin{eqnarray}
\label{pair}
\psi_{ij}^P{(x_i,x_j)}=\mu(x_i,x_j){\sum_{m=1}^{K}\omega^{(m)}\kappa^{(m)}{(f_i,f_j)}},
\end{eqnarray}
\end{small}
here $\mu$ is a label compatibility function, with Potts model $\mu(x_i,x_j) = 1_{[x_i\neq x_j]}$. $\kappa{(f_i,f_j)}= {\sum_{m=1}^{K}\omega^{(m)}\kappa^{(m)}{(f_i,f_j)}}$ is a linear combination of Gaussian kernels. $f_i$ is the feature vector from pixel $i$. For multi-class image labelling, DenseCRF~\cite{Krahenbuhl2012tu} uses contrast-sensitive two-kernel potentials defined in terms of the color and position vector $I, P$:
\begin{small}
\begin{eqnarray}
\label{ap_sm}
\kappa(f_i,f_j)=\omega^{(1)} \underbrace{\text{exp}{(-\frac{{|P_i-P_j|}^2}{2{\theta_{\alpha}}^{2}}-\frac{{|I_i-I_j|}^2}{2{\theta_{\beta}}^{2}})}}_{\text{appearance kernel}}  
\\\nonumber+\omega^{(2)} \underbrace{\text{exp}{(-\frac{{|P_i-P_j|}^2}{2{\theta_{\gamma}}^{2}})}}_{\text{smoothness kernel}} 
\end{eqnarray}
\end{small}

\textbf{Region-based Higher-order Potentials.} Region-based potentials mainly use $P^N$ Potts model. In Dense+Potts~\cite{vineet2014filter}, H-CRF-RNN~\cite{arnab2016higher}, each clique is a is superpixel which consist of all the pixels inside. The higher-order potentials with $P^n\text{-}Potts$ model type energy~\cite{kohli2007p3} are shown in Eq.~\ref{term_sp}.
\begin{small}
\begin{eqnarray}
\label{term_sp}
\psi_s^{SP}{(X_s=x_s)}=\begin{cases}
\omega_{Low}(l), \text{ if all } x_s^{i} = l, \cr \omega_{High}, \text{otherwise}.
\end{cases}
\end{eqnarray}
\end{small}
where $x_s$ corresponds to all the pixels within the $s^{th}$ superpixel, $\omega_{Low}(l) < \omega_{High}$ for all $l$. Costs $\omega_{Low}(l)$ and $\omega_{High}$ are weights need to be learned.  

The Robust $P^N$ model is defined as 
\begin{small}
\begin{eqnarray}
\label{robust_higher_order}
\psi_s^{SP}{(X_s=x_s)}=\min_{l\in\EuScript{L}} (\gamma_c^{max}, r_c^{l}+k_c^{l}N_c^l),
\end{eqnarray}
\end{small}
where $N_c^l = \sum_{i\in c} \delta({x_i\neq l})$ is the number of inconsistent pixels in the clique $c$ with the label $l$. 

\section{The Proposed Methods}
\label{sec:proposed}
First, in Section~\ref{observation}, we present the two important observations related to higher-order potentials and the pairwise potentials. Then, we describe the first layer of our approximation approach which is SaI in Section~\ref{first_layer}. In Section~\ref{sec: multilayer CRF}, we explain how the multi-layer approximation approach SM-CRF works. Section~\ref{application} provides the details of the application of SaI and SM-CRF on DenseCRF~\cite{russell2012exact} and CRF-RNN~\cite{Zheng2015cu}.

\subsection{Important Observations}
\label{observation}
We make the following two important observations on the region-based higher-order potentials and the pairwise potentials:

(1) This higher order potential shown in Eq.~\ref{robust_higher_order} is then formulated as a minimization of a secondary pairwise CRFs between an auxiliary variable $y_c$ and the pixel within the higher order clique. $y_c$ takes value from the extended label set $\EuScript{L}^E = \EuScript{L} \cup \EuScript{L_F}$. Here the observed sequence data is all the pixels in the higher order clique. 
\begin{align}
\psi_s^{SP}{(X_s)}=\min_{y_c \in \EuScript{L}^E} (\phi_c(y_c) + \sum_{i\in c}\phi_c(y_c, x_i))~\cite{kohli2009robust, russell2009associative}  
\end{align}
This conversion of higher order potential makes us believe that the higher-order regulation eventually ``flows'' down onto pairwise cliques. 

(2) When the color brightness is used as a feature vector in the pairwise potentials the value of potentials is inversely proportional to the absolute color brightness difference of two pixels. This color-sensitive feature makes the pairwise potentials obtain the maximum value at $I_i = I_j$ as shown in Eq.~\ref{max_pair_potential}. This encourages pixels with high resemblance to take the same label and vice versa. 
\begin{small}
\begin{eqnarray}
\label{max_pair_potential}
{(I_i,I_j)}^\ast = \arg\max_{I_i,I_j\in\EuScript{L}^{\EuScript{N}}}\psi_{ij}^P{(x_i,x_j; \bm{\theta}, \EuScript{F}\fgebackslash I)}\\\nonumber
{(I_i,I_j)}^\ast = (I_i = I_j)
\end{eqnarray}
\end{small}
$\EuScript{F}\fgebackslash I$ represents all feature vectors but $I$, e.g. $P$. $\bm{\theta}$ denotes all the hyperparameters and weights of this model. 
\setlength{\intextsep}{2pt}
\begin{figure}
\centering 
             \begin{subfigure}[b]{0.49\linewidth}
                \includegraphics[width=0.98\columnwidth, height = 1.8cm]{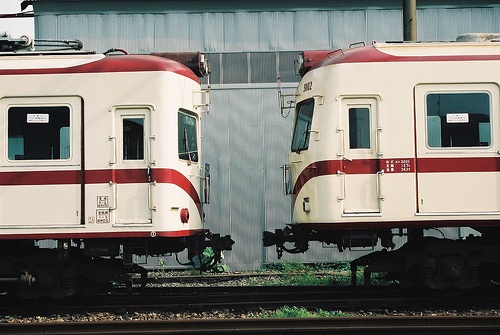}
                \caption{Original Image}
        \end{subfigure}%
             \begin{subfigure}[b]{0.49\linewidth}
                \includegraphics[width=.98\columnwidth, height=1.8cm]{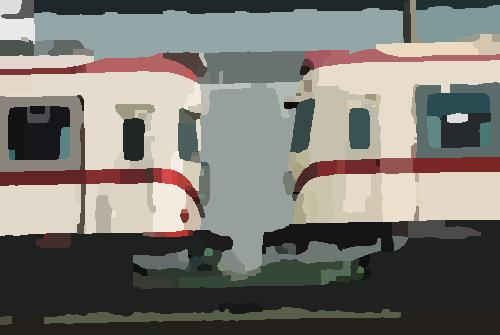}
                \caption{$\bm{D}_s$ with BGPS~\cite{li2012segmentation}}
        \end{subfigure}%
         \caption{The examples of the segmented image $\bm{D}_s$ from BGPS~\cite{li2012segmentation}.}
                 \label{fig:fig_sp}
\end{figure}
\subsection{The First Layer: Segmentation as Input CRF (SaI)}
\label{first_layer}
Inspired by observation one, we proposed Segmentation as Input (SaI) method to form the higher-order cues directly on pairwise potentials to resolve the computational complexity faced by traditional higher-order CRFs in the inference and learning, Suppose we form a pairwise graph on the segmentation of a given image, for the pairwise clique: if it exists inside one superpixel, which we name such edge as intra-edge and the potential as intra-potential, $\psi_{intra}(x_i,x_j)$,and if it exists across two different superpixels, which we name such edge as extra-edge and the potential as extra-potential, $\psi_{extra}(x_i,x_j)$. Then it is reasonable to have the following equation:
\begin{align}
\label{relation}
\psi_{intra}(x_i,x_j)\geq \psi_{extra}(x_i,x_j),\delta(x_i,x_j) = 1, 
\end{align}
so that label consistency in segments is enforced. SaI is proposed to satisfy this scheme to approximate both the higher-order and pairwise potentials.

\textbf{Segmentation as Input.}   In SaI, we pre-process the original RGB images from the dataset with unsupervised segmentation, in our paper, we used mean-shift~\cite{comaniciu2002mean} and BGPS~\cite{li2012segmentation} as shown in Fig.~\ref{fig:fig_sp}.  We use $s_i$ and $s_j$ as the segment indexes of pixel $i$ and $j$, respectively. Then we store a segmented image wherein each pixel $i$ takes the average RGB value $C_{s_i}$ of the superpixel that it belongs to. We noted such segmented image as $\bm{D}_s$. The SaI based pairwise potential is denoted as ${\psi^{SaI}_{ij}}{(x_i,x_j; \bm{D}_s)}$. For example, if we formulate our SaI potential as $\psi^{SaI}_{ij}{(x_i,x_j)}=\mu(x_i,x_j)({\theta_p+\theta_v\text{exp}(-\theta_\beta|C_{s_i}-C_{s_j}|^2)})$, we have our intra potentials as $\psi^{SaI}_{intra}(x_i,x_j) = \mu(x_i,x_j)(\theta_p+\theta_v), s_i = s_j$. According to Observation two the intra-potential is the maximum which makes Eq.~\ref{relation} satisfied. 

\textbf{Relation to Robust $P^N$ Potts Potential.} Inside the pairwise graph that built on $\bm{X}$ and observed on $\bm{D}_s$, pairwise edges can be divided into two groups, the intra-edges denoted as $\EuScript{E}_{intra}$ and the extra-edges as $\EuScript{E}_{extra}$, which we have $\EuScript{E} =\EuScript{E}_{intra} \cup \EuScript{E}_{extra} $. 
The sum of all intra-potentials are further decomposed as follows: 
\begin{eqnarray}
\label{sum_intra_potentials}
\sum_{(i,j)\in\EuScript{E}_{intra}}{\psi^{SaI}{(x_i,x_j)}} &= \sum_{c\in\EuScript{S}}\sum_{i,j\in c}{\psi^{SaI}_{intra}(x_i,x_j)}, \\\nonumber
&=\sum_{c\in\EuScript{S}} (N_c(X_s) (\theta_p+\theta_v))
\end{eqnarray}
wherein $N_c(X_s)$ is the number of pairwise edges in the clique $c$ that take inconsistent labels. This equation shows the equivalency of the sum of intra-potentials in the pairwise graph as the Robust $P^N$ Potts potential in Eq.~\ref{robust_higher_order}. The sum of all extra-potentials functions as a regulator between segments. 

SaI based pairwise CRF takes the following Gibbs energy function:
\begin{small}
\begin{eqnarray}
\label{pho-crf}
E(\bm{X}|\bm{D}_s)&= 
{\sum_{i\in \EuScript{V}}{\psi_i}^U{(x_i)}+\sum_{(i,j)\in\EuScript{E}}{\psi^{SaI}_{ij}}}{(x_i,x_j)}
\end{eqnarray}
\end{small}
The benefit of this model is that the computational complexity is same as that of the original pairwise CRFs of both the learning and inference processes. However, SaI does slightly suffer from the pairwise regulation loss. This is due to: when $s_i \neq s_j$, we have $|C_{s_i} - C_{s_j}| \cong |I_i - I_j|$ compared with $\bm{D}$, a gap of the left and the right sides of the equation exists. This gap potentially depends on the resulting segments, the smaller each superpixel is, the more identical the two sides of the equation are. The proposed SaI-based Multi-layer CRFs Framework (SM-CRF) incorporates another pairwise layer after SaI to compensate the potential drawback of SaI. 
\subsection{SaI-based Multi-layer CRFs Framework}
\label{sec: multilayer CRF}
\setlength{\intextsep}{2pt}
\begin{figure}[!htb]
\centering 
    \includegraphics[width=1\linewidth,height = 3.5cm]{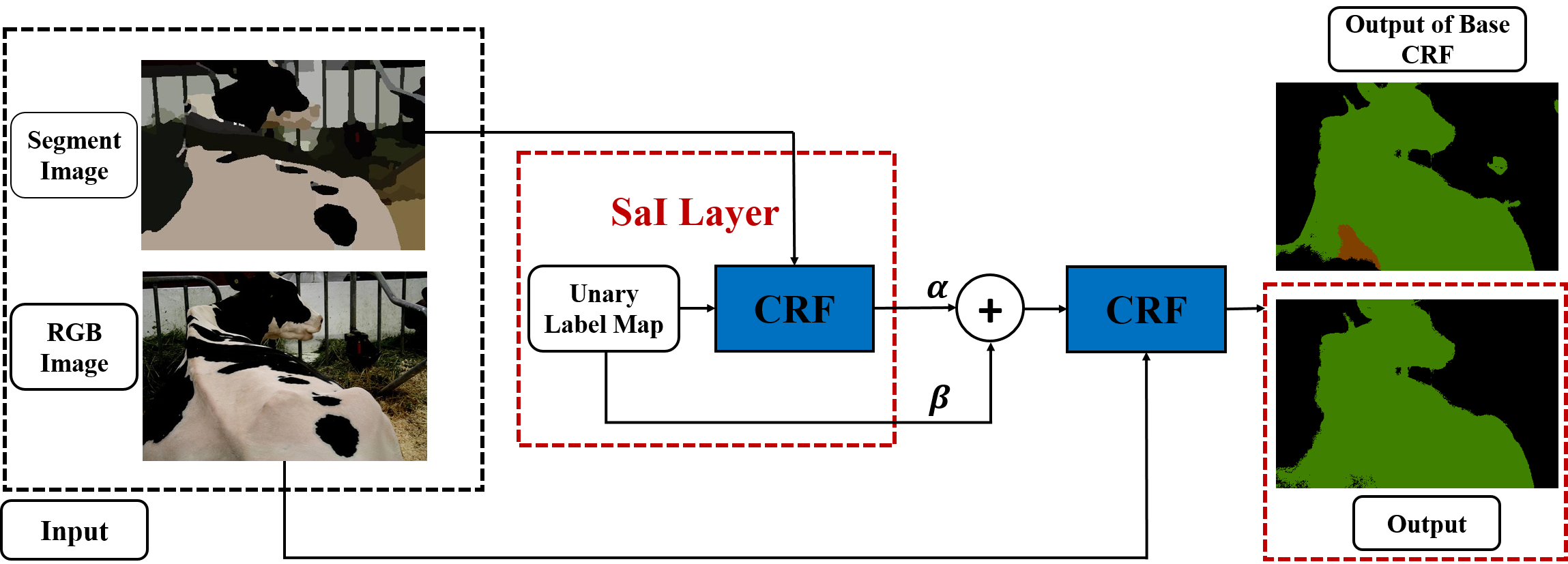}
\captionof{figure}{SM-CRF Framework. In the middle red box is the first layer of SaI. We observe performance improvement comparing of the output of SM-CRF noted in the red box on the right side with the output of the original CRF.}
 \label{fig:fig_frame}
\end{figure}

The framework of SaI-based Multi-layer CRFs (SM-CRF) is shown in Fig.~\ref{fig:fig_frame}. SM-CRF is a two layer pairwise CRF wherein the first layer is SaI which can be looked upon as a higher-order potential term and the second layer is a pairwise CRF, which takes the original RGB image as input instead. Each layer's Gibbs energy function is shown in Eq.~\ref{pho-crf} and Eq.~\ref{ge_layer2}. The initial unary probability label map is $U(x)$, the output unary map from SaI layer is denoted as $U_{SaI}^{(1)}(x)$, and the unary map as input to the second layer is denoted as $U^{(1)}(x)$. We set $U^{(1)}(x)= \alpha*U_{SaI}^{(1)}(x)+\beta*U(x)$, $\alpha+\beta=1$. This is trained with grid search. 
\begin{small}
\begin{eqnarray}
\label{ge_layer2}
E(\bm{X}|\bm{D})^{(2)}=\sum_{i \in \EuScript{V}}\psi^{U^{(1)}}{(x_i)}+\sum_{(i,j)\in\EuScript{E}}\psi^P{(x_i,x_j)}
\end{eqnarray}
\end{small}
This multiple layer structure helps the first SaI layer to incorporate more efficient pairwise regulations. The result is higher accuracy and better performance at preserving the details of the boundaries.

\subsection{The Application of SaI and SM-CRF on Fully Connected CRF Models}
\label{application}
We focus on two basic models: DenseCRF~\cite{russell2012exact, krahenbuhl2013parameter} and CRF-RNN~\cite{Zheng2015cu} which represents simple Potts model where $\mu(x_i,x_j)=1_{[x_i\neq x_j]}$ being used as post-processing tool after classifier and more advanced model wherein CRF is formulated as a stack of Recurrent Neural Networks which is trained end-to-end with the classifier CNN, repectively. In DenseCRF, we use setting $\omega^{(m)},\mu \in \mathcal{R}^1$, while in CRF-RNN $\omega^{(m)},\mu \in \mathcal{R}^{\EuScript{L}\times\EuScript{L}}$ and are learned on large dataset. To update these two basic models to incorporate higher-order cues, we set our SaI-based Pairwise potentials to be the same as the formulation of DenseCRF shown in Eq.~\ref{pair} and \ref{ap_sm}. In the case that we do not have a pre-trained basic models, (1) the learning and inference of SaI will be the same as the basic models, and (2) the learning of SM-CRF is in a layer-by-layer manner. First to train SaI and then fix the parameters of the first layer to be the trained optimum values to train parameters involved with adding the second layer. However, if there is already a pre-trained pairwise model that one can just perform a simple upgrade to obtain a superpixel-based higher-order CRF. 

\textbf{Upgrade Pre-trained Pairwise Model to SaI model. } As shown in DenseCRF~\cite{russell2012exact, krahenbuhl2013parameter}, $\omega^{(1) }$ and $\theta_\alpha$ are the most important hyperparameters or weights. Thus, to keep the new model simple but effective, in SaI the pairwise potentials reuse all the other hyperparamters or weights except $\omega^{(1) }$ and $\theta_\alpha$, which include compatibility function $\mu$, $\omega^{(2)}$, $\theta_\beta$, and $\theta_\gamma$. 

\textbf{For DenseCRF.} If there is a trained pairwise models, to upgrade this model to incorporate superpixel-level cues, we just need to do simple grid search for the two most important parameters $\omega^{(1)}$ and $\theta_\alpha$ in model SaI.

\textbf{For CRF-RNN.} Because in CRF-RNN, $\omega^{(1)}$ is a $21\times 21$ matrix that learned on large training dataset. Thus, we propose the relation $\omega_{SaI}^{(1)} = r \omega^{(1)}$, $r\in(0,1]$ to effectively reuse this parameter. Therefore, we can still use simple grid search of $r$ and $\theta_\alpha$ to upgrade this pairwise model to SaI model. With the upgrade, the new SaI based pairwise potential takes the following form:
\begin{eqnarray}
\psi^{SaI}_{i,j}(x_i,x_j)&=\{r\omega^{(1)}{\exp{(-\frac{{|P_i-P_j|}^2}{2{\theta_{{\alpha}_{SaI}}}^{2}}-\frac{{|C_{s_i}-C_{s_j}|}^2}{2{\theta_{{\beta}}}^{2}})}}\\\nonumber
&+ \omega^{(2)} \exp(-\frac{{|P_i-P_j|}^2}{2\theta_\gamma ^2})\}\mu(x_i,x_j)
\end{eqnarray}

\textbf{Upgrade Pre-trained Pairwise Model to SM-CRF Model.} We need to train SM-CRF layer by layer. At first, we perform the first round of grid search to obtain $\omega^{(1)}$ and $\theta_\alpha$  for the first SaI layer by setting setting $\alpha = 0.5, \beta = 0.5$. Next, we obtain one more hyperparameter $\alpha$ ($\beta = 1-\alpha$) which connects the initial unary map and the unary output map of SaI layer.
\section{Experiments}
\label{sec:result}

\subsection{Experimental Setup}
\label{setup}
In our experiments, we evaluated our methods on two public datasets:  MSRC-$21$ and PASCAL VOC $2012$~\cite{hariharan2014simultaneous}. In MSRC-$21$, there are a total of $591$ RGB images with coarse labeled ground truth of $21$ object classes~\cite{shotton2009textonboost}, which is called Standard Ground Truth (SGT). Such coarse ground truth with unlabeled regions around the object boundaries makes it difficult to quantitatively evaluate the algorithm's performance, since the algorithm crave for the pixel-level accuracy. Therefore, we used a subset from MSRC-$21$ of $92$ images with pixel-level accurate labelling, which is noted as Accurate Ground Truth (AGT). PASCAL VOC $2012$  has a total of $11,685$ images with a total of $20$ object classes in the training and validation set. In our experiment, we evaluated the performance on a reduced validation set (RVS) as used in \cite{Zheng2015cu, arnab2016higher}, which includes $346$ images. Three metrics are used for evaluating the proposed methods, global pixel accuracy (Global), averaged pixel accuracy (Average), and mean intersection over union (MeanIoU). 
\subsection{Evaluation with DenseCRF on MSRC-$21$}
\label{experiment_1}
First, we segmented both SGT and AGT with mean-shift segmentation with hyper-parameters set as $(h_s, h_r) = (7, 6.5)$. The output from the segmentation is noted $\bm{D}_s$. Then we split AGT into half as training set and half as testing set. All the models were tested on the testing set in AGT. We use the same unary potentials that were used in the implementation of DenseCRF~\cite{Krahenbuhl2012tu} which was derived from TextonBoost~\cite{shotton2009textonboost,russell2009associative}. Next, we generated two models based on DenseCRF. 1) SaI, which is the same as DenseCRF, except that our observation is $\bm{D}_s$ instead of $\bm{D}$. 2) SM-CRF; consists of SaI followed by a second layer which is a pairwise CRF with $\bm{D}$. Here we set the comparatively less important hyperparameters to be $\theta_\beta = 13, \theta_\gamma = 3$, and $\omega^{(2)} = 3$.

\begin{figure}
    \centering
    \includegraphics[width=0.325\linewidth, height = 2.5cm]{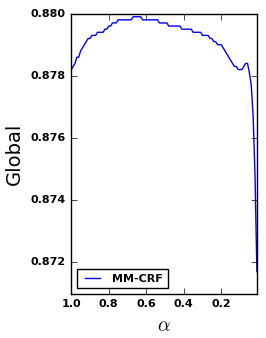}
    \includegraphics[width=0.325\linewidth, height = 2.5cm]{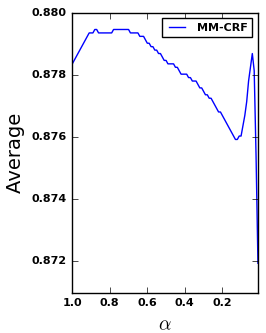}
    \includegraphics[width=0.325\linewidth, height = 2.5cm]{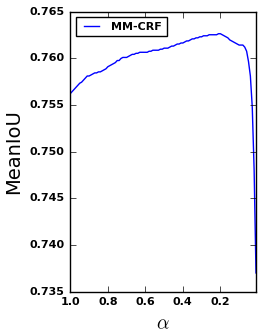}
    \caption{Grid Search of parameter $\alpha$ for SM-CRF on MSRC-$21$, $\beta = 1-\alpha$. When $\alpha = 1$, the SM-CRF has degraded into a single layer of DenseCRF.}
    \label{fig:train_a}
                \includegraphics[width=0.19\linewidth]{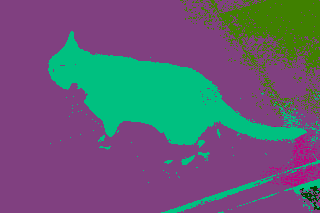}
                \includegraphics[width=0.19\linewidth]{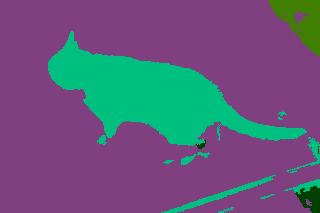}
                \includegraphics[width=0.19\linewidth]{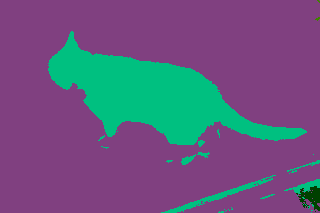}
                \includegraphics[width=0.19\linewidth]{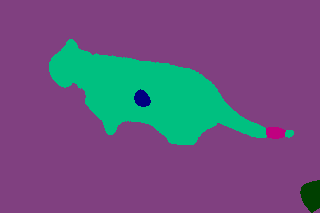}
                \includegraphics[width=0.19\linewidth]{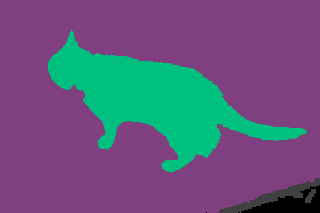}

                \includegraphics[width=0.19\linewidth]{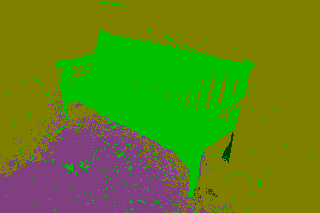}
                \includegraphics[width=0.19\linewidth]{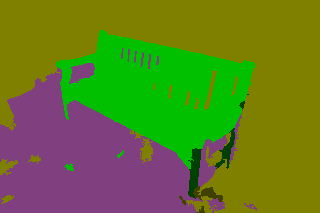}
                \includegraphics[width=0.19\linewidth]{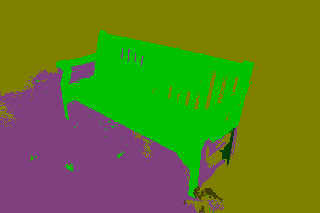}
                \includegraphics[width=0.19\linewidth]{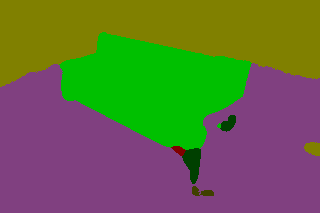}
                \includegraphics[width=0.19\linewidth]{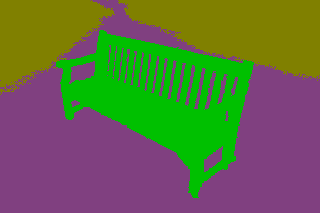}

                \includegraphics[width=0.19\linewidth]{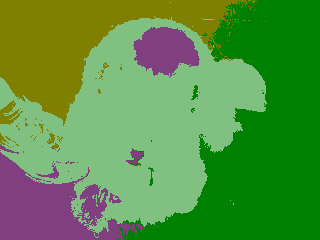}
                \includegraphics[width=0.19\linewidth]{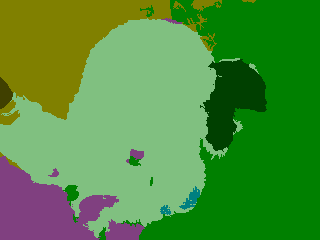}
                \includegraphics[width=0.19\linewidth]{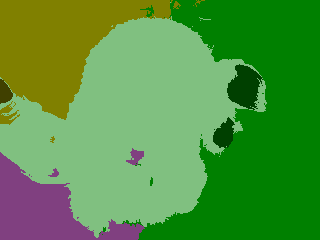}
                \includegraphics[width=0.19\linewidth]{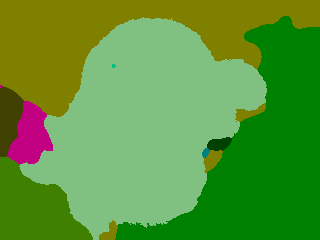}
                \includegraphics[width=0.19\linewidth]{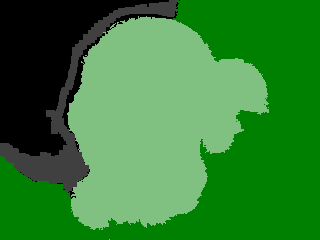}

                \includegraphics[width=0.19\linewidth]{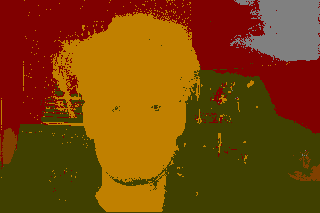}
                \includegraphics[width=0.19\linewidth]{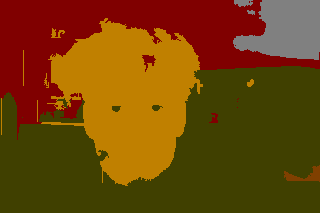}
                \includegraphics[width=0.19\linewidth]{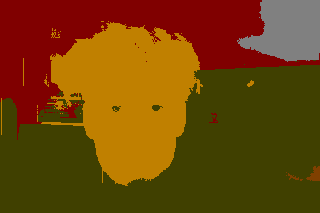}
                \includegraphics[width=0.19\linewidth]{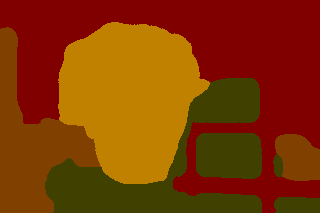}
                \includegraphics[width=0.19\linewidth]{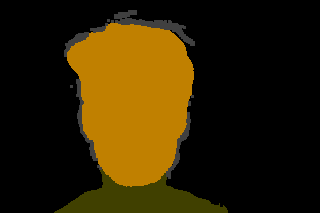}

                \includegraphics[width=0.19\linewidth]{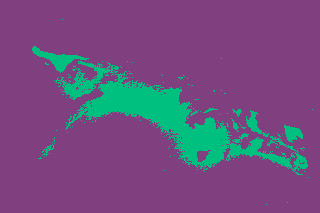}
                \includegraphics[width=0.19\linewidth]{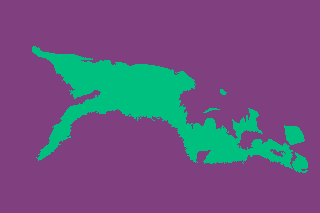}
                \includegraphics[width=0.19\linewidth]{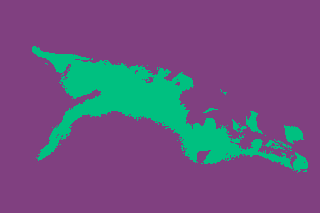}
                 \includegraphics[width=0.19\linewidth]{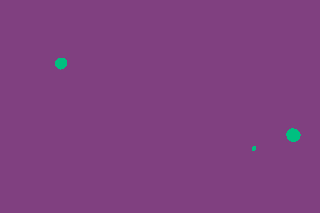}
                \includegraphics[width=0.19\linewidth]{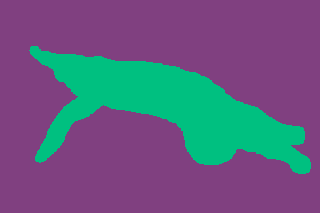}

                \includegraphics[width=0.19\linewidth]{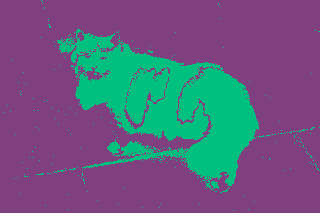}
                \includegraphics[width=0.19\linewidth]{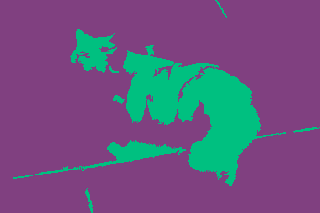}
                \includegraphics[width=0.19\linewidth]{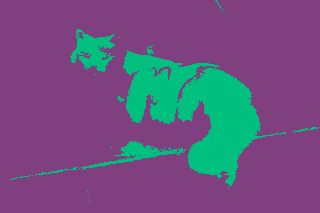}
                \includegraphics[width=0.19\linewidth]{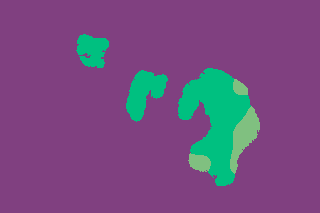}
                \includegraphics[width=0.19\linewidth]{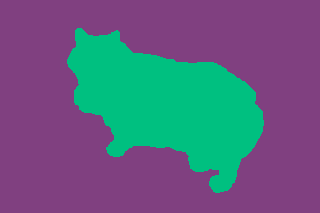}

                \includegraphics[width=0.19\linewidth]{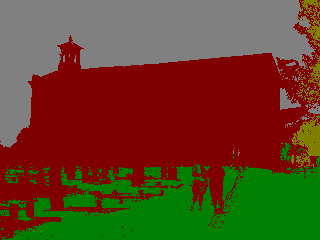}
                \includegraphics[width=0.19\linewidth]{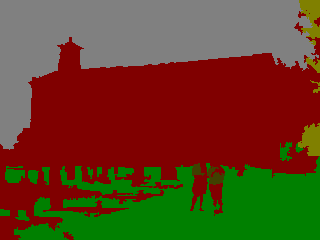}
                \includegraphics[width=0.19\linewidth]{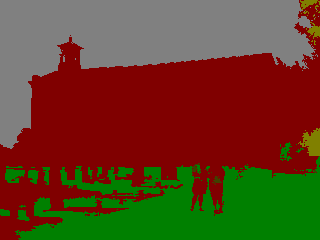}
                \includegraphics[width=0.19\linewidth]{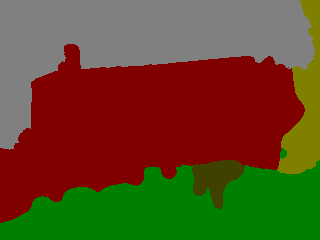}
                \includegraphics[width=0.19\linewidth]{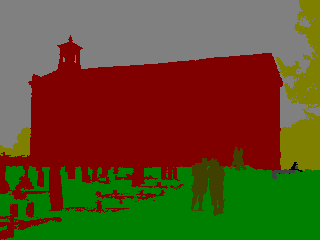}
        \caption{Examples of qualitative results on MSRC-$21$ dataset. From left to right, it is DenseCRF, SaI-CRF, SM-CRF, Dense$+$Potts, and GT. Row one to row five are successful examples and the last two rows are the failure cases.}
        \label{fig:msrc}
\end{figure}

\textbf{The Effectiveness of Simple Grid Search.} We performed grid search for DenseCRF and SaI with two sets of training. First, we set $\theta_\alpha = 80$ for both models to train $\omega^{(1)}$ which is the weight of the appearance kernels, we obtained $\omega^{(1)} = 6$ for DenseCRF and $10$ for SaI. Then, we fixed $\omega^{(1)}$ and trained to find $\theta_\alpha$, which was $48$ for DenseCRF and $66$ for the SaI, respectively. For SM-CRF, an additional set of grid search is required for $\alpha, \beta$ and the training process is shown in Fig.~\ref{fig:train_a}. The trained parameter was $\alpha = 0.63, \beta = 0.37$. For the comparative model Dense$+$Potts~\cite{vineet2014filter}, it needs to be trained on a large dataset to obtain workable higher-order potentials. Here, we use SGT minus the testing set from AGT as its training set. After the higher-order potential is trained, we did grid search to train its two hyperparameters. The testing result of each model is shown in Table~\ref{tab:msrc_precession}. Our models' performance on the testing set demonstrates the effectiveness of learning with simple grid search without another set of training for the higher-order potentials needed in Dense+Potts model. 

\setlength{\intextsep}{2pt}
\begin{table}[!ht] 
\small
\centering
\noindent\captionof{table}{ Quantitative results on MSRC-$21$ dataset.}
 \noindent \begin{tabular}{|p{0.23\columnwidth}|p{0.15\columnwidth}| p{0.15\columnwidth}| p{0.15\columnwidth}| p{0.12\columnwidth}|}
  \hline
    & \multicolumn{3}{|c|}{Accurate Ground Truth}& Time\\ \hline\hline
   & Global&Average&IOU& \\ \hline
  Unary  &$82.33$&$83.30$&$63.18$& $N/A$ \\ \hline
DenseCRF~\cite{russell2012exact}  & $86.63$&$86.29$&$71.80$&$0.16s$ \\ \hline
Dense$+$Potts~\cite{vineet2014filter} & $85.76$ &$84.41$& $72.73$ &$1.14s$\\ \hline\hline
\textbf{SaI}& ${87.57}$&${87.25}$&$74.62$&$0.16s$\\ \hline\hline
\textbf{SM-CRF} & $\mathbf{87.99}$&$\mathbf{87.93}$&$\mathbf{76.06}$&$0.42s$\\ \hline
\end{tabular}
  \label{tab:msrc_precession}
\end{table}

From Table~\ref{tab:msrc_precession}, the accuracy improvement we gained for SaI and SM-CRF compared with DenseCRF is ($0.94\%$, $1.36\%$) on Global, ($0.96\%$, $1.64\%$) on Average, and ($2.82\%$, $4.26\%$) on MeanIoU, respectively. The model SaI consumed the same amount of time as DenseCRF, and for SM-CRF, the time is slightly more than double of DenseCRF. The proposed scheme decreased the error rate by as much as $16.3\%$. While, Dense$+$Potts gained an accuracy equivalent to DenseCRF. This may be due to the trained higher-order term, which can recommend a new label based on its dependent training. This can be both helpful and harmful. When the predicted label is right, it can be an advantage as row two in Fig.~\ref{fig:msrc} the ``grass'' marked by purple color is more complete than any of the other models. However, this new label might be a wrong one, especially  when the initial unary prediction is already of high accuracy, this in turn can make the accuracy worse. Row three and row five in Fig.~\ref{fig:msrc} demonstrate such defect, where we observed that the ``cat'' totally disappeared. Also, we noticed that Dense$+$Potts performed worse than our models in preserving sharp boundaries of objects.  
\setlength{\intextsep}{2pt}
\begin{table}
\small
\centering
\noindent\captionof{table}{ Performance comparison of SaI-RNN with CRF-RNN on different setting of parameters.}
  \noindent \begin{tabular}{|p{0.15\columnwidth}|p{0.25\columnwidth}| p{0.12\columnwidth}| p{0.12\columnwidth}| p{0.12\columnwidth}|}
    \hline
  	& CRF-RNN~\cite{Zheng2015cu} & SaI-$1$& SaI-$2$ & SaI-$3$ \\ 
	\hline\hline
parameters &$\mu_{t}, \omega_{t}^{(m)}$ & $\mu, \omega^{(m)}$ &$\mu, \omega_{t}^{(m)}$&$\mu_t, \omega_{t}^{(m)}$ \\ \hline
MeanIoU &$72.9$&  $71.9$&$73.3$&$\mathbf{73.5}$ \\
 \hline
  \end{tabular}
   \label{tab:firstlayer}
\begin{small}
\centering
\noindent\captionof{table}{ Performance comparison of our methods with CRF-RNN~\cite{Zheng2015cu} and H-CRF-RNN~\cite{arnab2016higher}. }
  \noindent \begin{tabular}{|p{0.25\linewidth}|p{0.11\linewidth}| p{0.11\linewidth}| p{0.12\linewidth}| p{0.2\linewidth}|}
    \hline
  	& Global  & Average & MeanIoU & Retrain \\ 
	\hline\hline
CRF-RNN~\cite{Zheng2015cu} &$91.9$& $81.2$&${72.9}$ & N/A\\\hline
H-CRF-RNN~\cite{arnab2016higher} &N/A&  N/A&$\mathbf{74.0}$ & Yes \\\hline
\textbf{SaI-RNN}& $91.55$ & $\mathbf{82.23}$&$73.54$& Grid Search\\\hline
\textbf{SM-CRF-RNN} &$\mathbf{91.94}$&  $81.51$ & $\mathbf{73.94}$ & Grid Search\\ 
 \hline
  \end{tabular}
   \label{tab:CRF-RNN-HO}
   \end{small}
\end{table}
\subsection{Evaluation with CRF-RNN on PASCAL VOC $2012$ }
\label{experiment_2}
First, we segmented our dataset RVS with BGPS segmentation of scale $15$ and saved the segmented image as $\bm{D}_s$. The weights of both CNN and CRF-RNN were tuned end-to-end in the implementation~\footnote{https://github.com/torrvision/crfasrnn}. We evaluated two models based on CRF-RNN~\cite{Zheng2015cu}: SaI-RNN, where we just changed the input to CRF-RNN from $\bm{D}$ to $\bm{D}_s$ and SM-CRF-RNN where we have two layers, the input of each layer is $\bm{D}_s$ and $\bm{D}$, respectively. 

\textbf{The Effectiveness of Parameters Reuse. } In SaI-RNN, we have three different settings to demonstrate the effectiveness of reusing the parameters that trained from the baseline CRF-RNN~\cite{Zheng2015cu}, including $\mu$ and $\omega^{(m)}$ ($\omega^{(1)}$ and $\omega^{(2)}$) which are $21\times 21$ matrices.  Here, we used $\mu_{t}, \omega_{t}^{(m)}$ to denote the end-to-end tuned parameters from CRF-RNN.  In the first setting, we named it \textit{SaI-$1$}, and here we did not use any end-to-end tuned parameters; in the second setting, we only reused the $\omega_{t}$, named \textit{SaI-$2$}; and in the third model, we reused both $\mu_{t}$ and $\omega_{t}^{(m)}$, and named it \textit{SaI-$3$}. In all these settings, $r$ was simply set to $1$. 

From the results compared with CRF-RNN as shown in Table~\ref{tab:firstlayer}, we note that by reusing both $\mu_{t}$ and $\omega_{t}$ we gained $0.6\%$ improvement in MeanIoU even without any training process. When we only reused the $\omega_{t}$, the improvement was $0.4\%$ instead. However, we obtained lower accuracy by not reusing any trained parameters and/or weights. This demonstrates the effectiveness and efficiency of SaI, which is boosted accuracy without either training or more time cost in the inference given a pre-trained pairwise CRF.

\textbf{The Effectiveness of Simple Grid Search.} 
In this set of experiments, both SaI-CRF-RNN and SM-CRF-RNN used $\mu_{t}$ and $\omega_{t}^{(m)}$ in the model. We used a subset of $200$ images from the validation set of PASCAL VOC $2012$ to conduct the grid search. First, we trained parameters $\theta_\alpha$ and $r$ for the first layer SaI-RNN, which we set $\theta_\alpha = 96, r=0.6$.  Next, we did another set of grid search for $\alpha$ ($\beta = 1-\alpha$), and the parameter we gained is $\alpha = 0.67, \beta = 0.33$. 
\setlength{\intextsep}{2pt}
\begin{figure}
\centering 
                \includegraphics[width=0.19\linewidth]{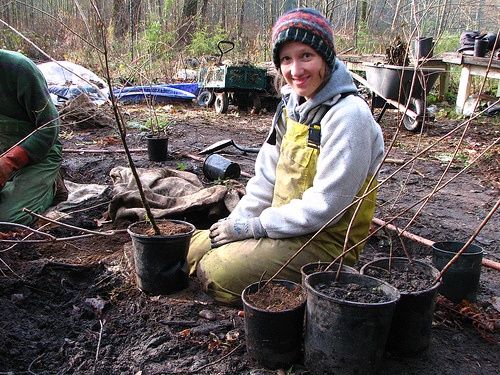}
                \includegraphics[width=0.19\linewidth]{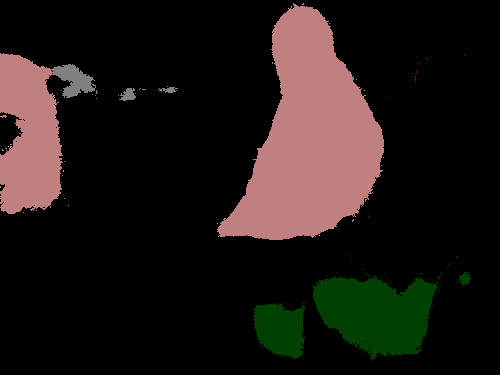}
                \includegraphics[width=.19\linewidth]{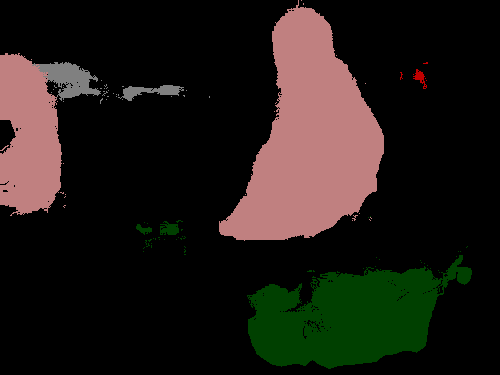}
                \includegraphics[width=.19\linewidth]{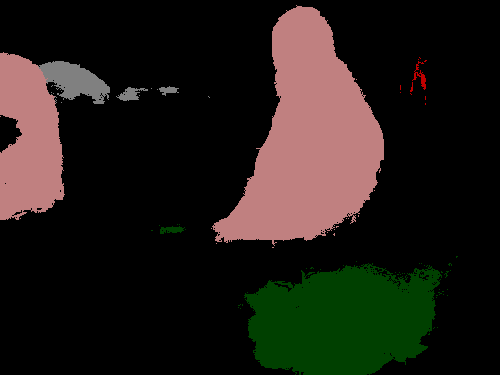}
                \includegraphics[width=.19\linewidth]{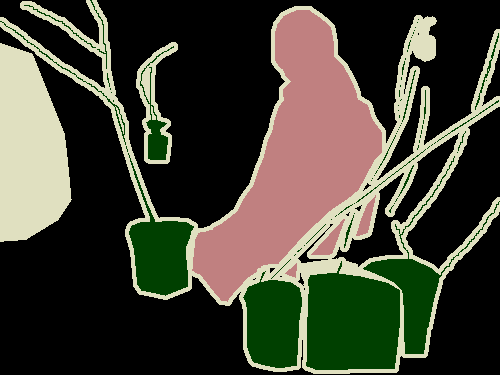}

                \includegraphics[width=0.19\linewidth]{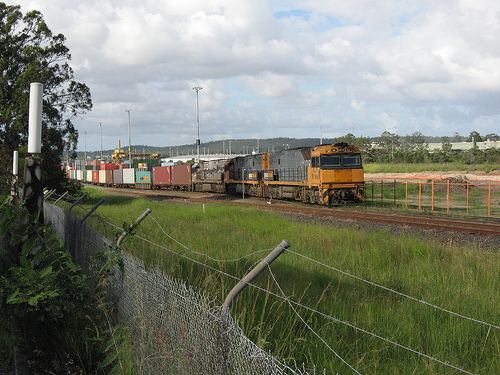}
                \includegraphics[width=0.19\linewidth]{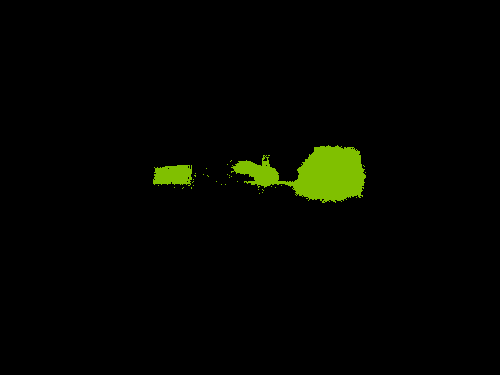}
                \includegraphics[width=.19\linewidth]{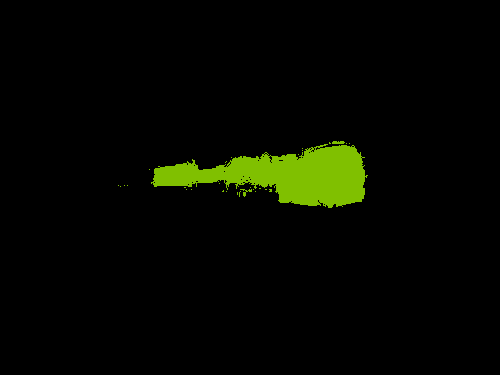}
                \includegraphics[width=.19\linewidth]{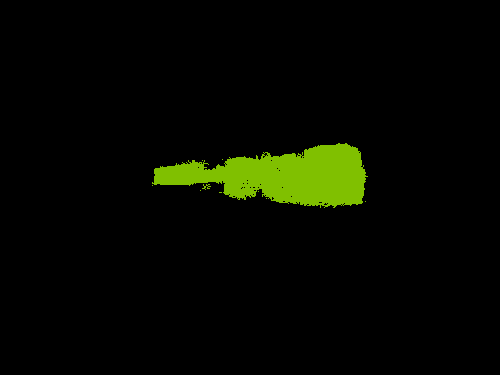}
                \includegraphics[width=.19\linewidth]{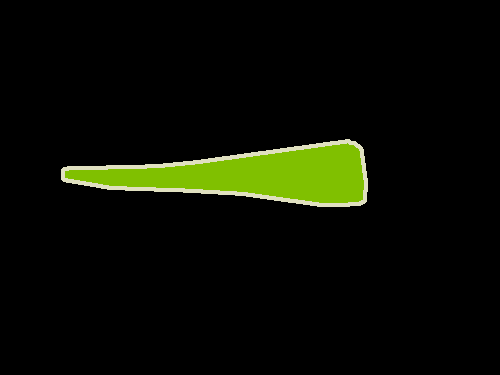}

                \includegraphics[width=0.19\linewidth]{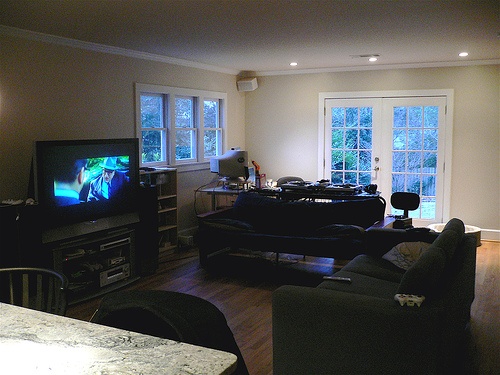}
                \includegraphics[width=0.19\linewidth]{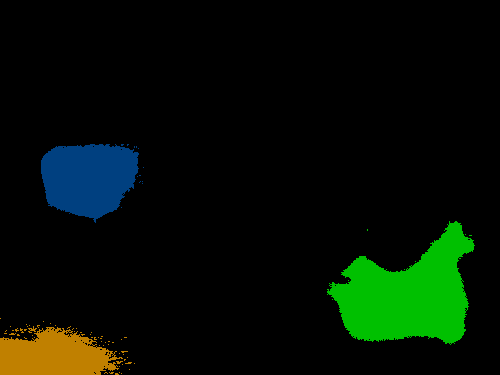}
                \includegraphics[width=.19\linewidth]{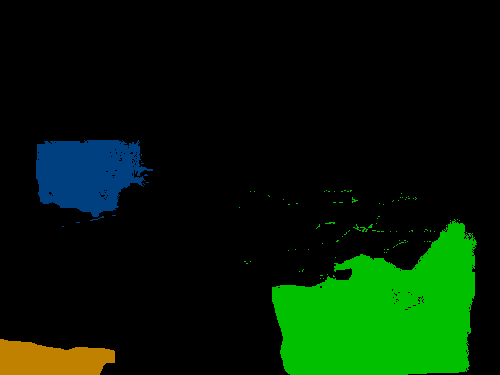}
                \includegraphics[width=.19\linewidth]{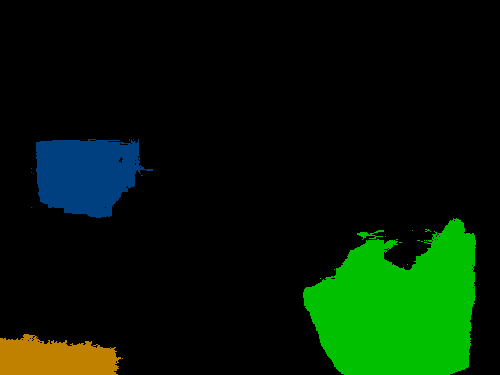}
                \includegraphics[width=.19\linewidth]{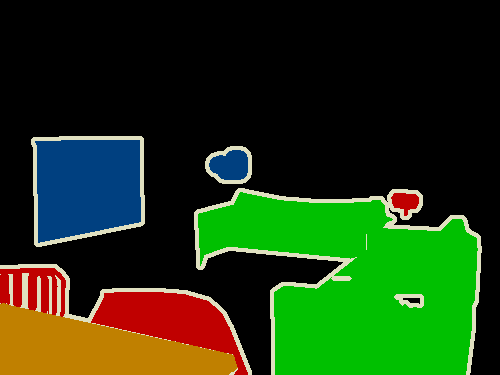}
        
                \includegraphics[width=0.19\linewidth]{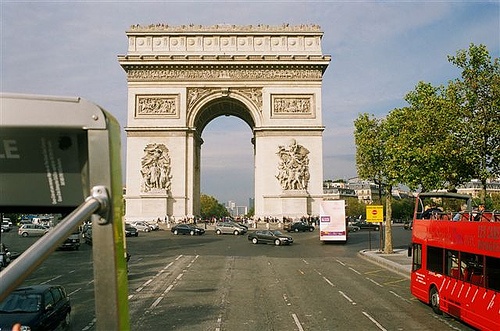}
                \includegraphics[width=0.19\linewidth]{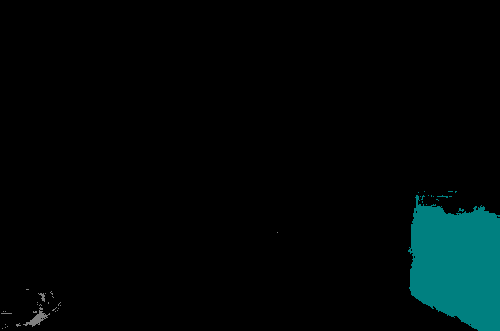}
                \includegraphics[width=.19\linewidth]{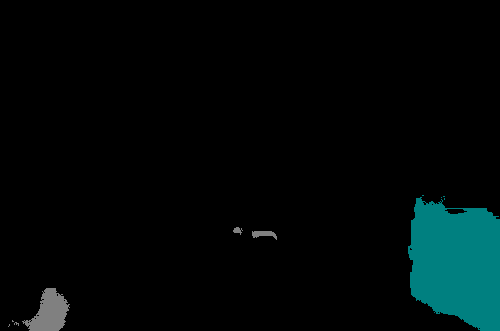}
                \includegraphics[width=.19\linewidth]{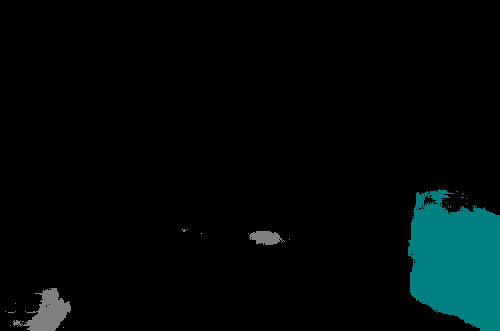}
                \includegraphics[width=.19\linewidth]{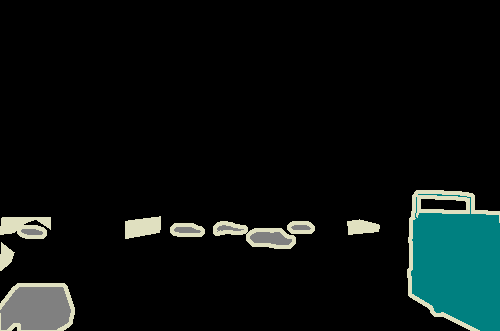}
        
                \includegraphics[width=0.19\linewidth]{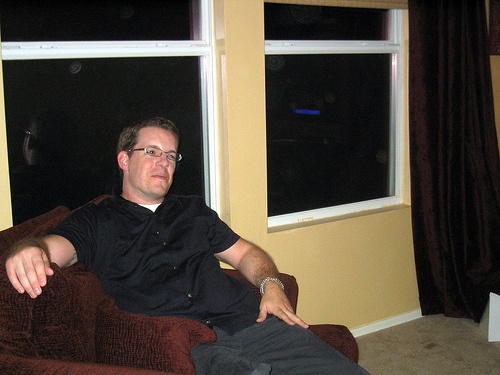}
                \includegraphics[width=0.19\linewidth]{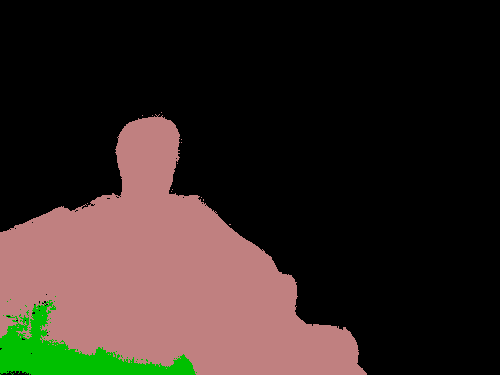}
                \includegraphics[width=.19\linewidth]{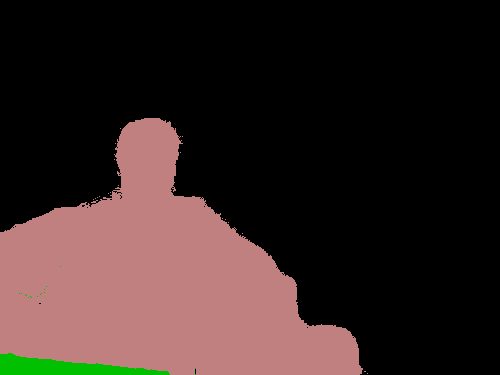}
                \includegraphics[width=.19\linewidth]{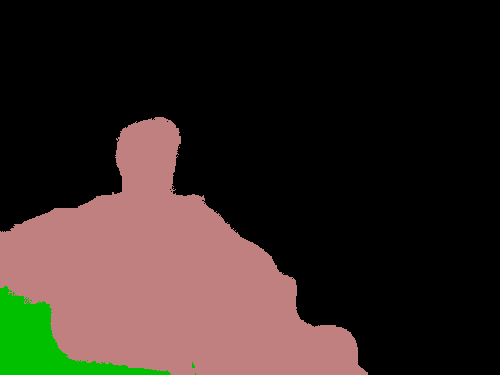}
                \includegraphics[width=.19\linewidth]{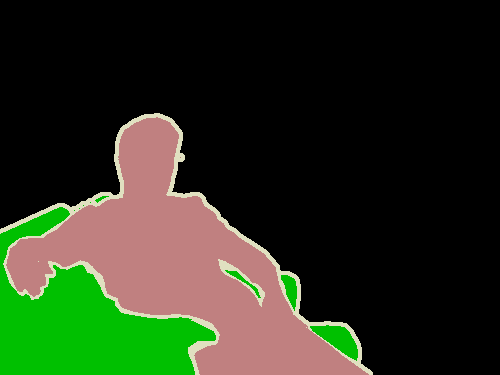}
                
                \includegraphics[width=0.19\linewidth]{fig/2007_008802.jpg}
                \includegraphics[width=0.19\linewidth]{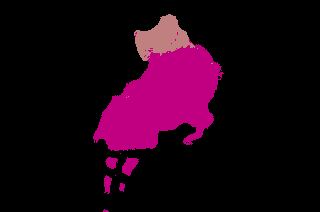}
                \includegraphics[width=.19\linewidth]{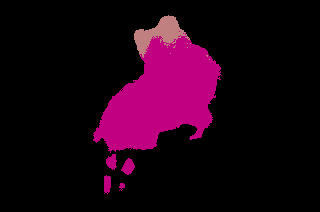}
                \includegraphics[width=.19\linewidth]{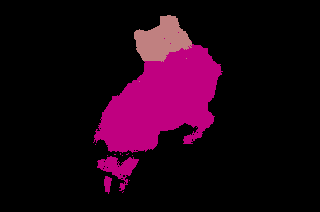}
                \includegraphics[width=.19\linewidth]{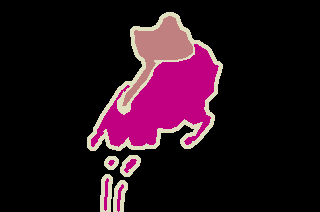}
        \caption{The comparison of segmentation results on PASCAL VOC $2012$ dataset. From left to right, it is Original Image, CRF-RNN, SaI-RNN, SM-CRF-RNN, and Ground Truth. Row one to row four are successful examples, and the last two rows show the failed cases. }
        \label{fig:fig_sucees_case}
\end{figure}

The quantitative results are shown in Table~\ref{tab:CRF-RNN-HO}, which indicate that we obtained $1\%$ MeanIoU accuracy improvement for SM-CRF-RNN, and for SaI-RNN, we gained $1\%$ Average accuracy improvement. The accuracy boost is equivalent to H-CRF-RNN~\cite{arnab2016higher}. The accuracy improvement we achieved through our models was on the condition that our models were either not trained or were trained only with a simple grid search with two more hyperparamters ( $r$  and $\alpha$). By contrast, H-CRF-RNN requires fine tuning on large dataset of more that $77,000$ images which consists of the PASCAL VOC $2012$ and Microsoft COCO dataset~\cite{lin2014microsoft}. The visual results are shown in Fig.~\ref{fig:fig_sucees_case}. The figures shows that our models perform well ``filling'' the incomplete predicted object. Also, the proposed scheme preserves sharp boundaries as well as getting rid of small spurious regions as shown in Fig.~\ref{fig:visual_result}. 

\section{Conclusions}
\label{sec:conclusion}
We presented SM-CRF, which is a two layer CRF framework that approximates traditional superpixel-based (region-based) higher-order CRFs. The first layer of SM-CRF, SaI, itself is a pairwise CRF, and it successfully integrates the superpixel-based higher-order cues into the color-sensitive pairwise potentials by feeding in the segmented image which encompassed the segmentation information. The second pairwise layer which takes the original RGB image as input compensates the pairwise regulation loss from the SaI and further boosts the accuracy. Our approximation approach achieves outstanding accuracy gain compared with the baselines, DenseCRF and CRF-RNN even without training or just by using a simple grid search while consuming the amount of time equivalent to those of the base models. When compared with other higher-order CRFs, DenseCRF$+$Potts and H-CRF-RNN, we either achieved better results or equivalent results with less time cost.

\bibliographystyle{IEEEbib}
{\bibliography{icme2018template}}

\end{document}